\documentclass[final]{cvpr}

\usepackage{times}
\usepackage{epsfig}
\usepackage{graphicx}
\usepackage{amsmath}
\usepackage{amssymb}

\usepackage[pagebackref=true,breaklinks=true,colorlinks,bookmarks=false]{hyperref}
\graphicspath{{figures/}}
\usepackage{booktabs}
\usepackage{makecell}
\usepackage{multirow}
\usepackage{amsmath}
\usepackage{amssymb}
\usepackage{epstopdf}

\def\cvprPaperID{1585} % *** Enter the CVPR Paper ID here
\def\confYear{CVPR 2021}

\begin{document}

%%%%%%%%% TITLE
\title{Rethinking BiSeNet For Real-time Semantic Segmentation}

\author{

    Mingyuan Fan\thanks{Equal contribution.},~
    Shenqi Lai\footnotemark[1],~
    Junshi Huang\thanks{Co-corresponding author.},~
    Xiaoming Wei\footnotemark[2],~
    Zhenhua Chai,\\
    Junfeng Luo,~
    Xiaolin Wei\\
    Meituan\\
    % \textsuperscript{\rm 1}Meituan\\
    \tt\small \{fanmingyuan, laishenqi, huangjunshi, weixiaoming, chaizhenhua, \\
    \tt\small luojunfeng, weixiaolin02\}@meituan.com
}
\maketitle

\pagestyle{empty}  % no page number for the second and the later pages
\thispagestyle{empty} % no page number for the first page
%%%%%%%%% ABSTRACT
\begin{abstract}
% In this work, we revisit BiSeNet~\cite{Yu2018BiSenet, Yu2020BiSeNetV2}, an efficient network architecture for real-time segmentation.

BiSeNet~\cite{Yu2018BiSenet, Yu2020BiSeNetV2} has been proved to be a popular two-stream network for real-time segmentation.
However, its principle of adding an extra path to encode spatial information is time-consuming, and the backbones borrowed from pretrained tasks, \textit{e.g.}, image classification, may be inefficient for image segmentation due to the deficiency of task-specific design.
To handle these problems, we propose a novel and efficient structure named Short-Term Dense Concatenate network (STDC network) by removing structure redundancy.  
Specifically, we gradually reduce the dimension of feature maps and use the aggregation of them for image representation, which forms the basic module of STDC network.
In the decoder, we propose a Detail Aggregation module by integrating the learning of spatial information into low-level layers in single-stream manner.
Finally, the low-level features and deep features are fused to predict the final segmentation results.
Extensive experiments on Cityscapes and CamVid dataset demonstrate the effectiveness of our method by achieving promising trade-off between segmentation accuracy and inference speed.
On Cityscapes, we achieve 71.9\% mIoU on the \textit{test} set with a speed of \textbf{250.4 FPS} on NVIDIA GTX 1080Ti,
which is \textbf{45.2\% faster} than the latest methods, and achieve \textbf{76.8\% mIoU} with 97.0 FPS while inferring on higher resolution images.
Code is available at \url{https://github.com/MichaelFan01/STDC-Seg}.
\end{abstract}
\vspace{-0.5cm}
%%%%%%%%% BODY TEXT
\section{Introduction}
% \noindent Semantic segmentation is a classic and fundamental topic in computer vision, 

\begin{figure}[t]
    \centering
    \includegraphics[width=1.0\columnwidth]{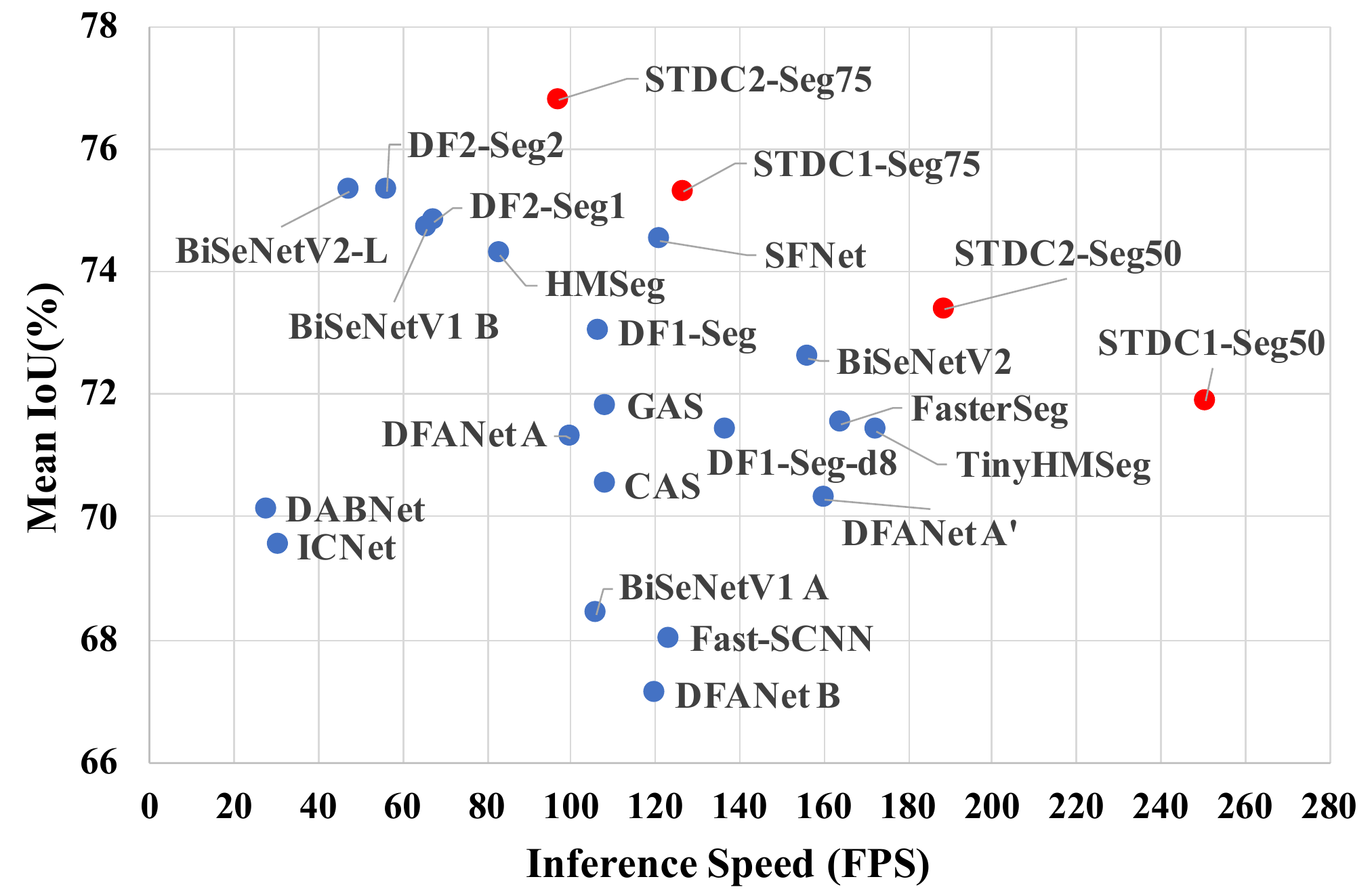} % Reduce the figure size so that it is slightly narrower than the column. Don't use precise values for figure width.This setup will avoid overfull boxes.
    \caption{Speed-Accuracy performance comparison on the Cityscapes \textit{test} set. 
    Our methods are presented in red dots while other methods are presented in blue dots.
    % The competitors include ICNet~\cite{zhao2018icnet}, DABNet~\cite{li2019dabnet}
    % , Fast-SCNN~\cite{poudel2019fastscnn}, BiSeNetV1~\cite{Yu2018BiSenet}, BiSeNetV2~\cite{Yu2020BiSeNetV2}
    % , GAS~\cite{Lin2020GAS}, CAS~\cite{zhang2019CAS}
    % , DFANet~\cite{li2019DFANet} and DFNet~\cite{Li2019DFNet}.
    Our approaches achieve state-of-the-art speed-accuracy trade-off.}
    \label{speed-accuract-comparison}
    \vspace{-0.7cm}
 \end{figure}
% To fulfill those demands, many researchers propose to design low-latency, high-efficiency CNN models with satisfactory segmentation accuracy.
% These real-time semantic segmentation methods have achieved promising performance on various benchmarks.
Semantic segmentation is a  classic and fundamental topic in computer vision, 
which aims to assign pixel-level labels in images.
The prosperity of deep learning greatly promotes the performance of semantic segmentation by making various breakthroughs \cite{li2019DFANet, Yu2020BiSeNetV2, Lin2020GAS, chen2020fasterseg},
coming with fast-growing demands in many applications, \textit{e.g.}, autonomous driving, video surveillance, robot sensing, and so on.
These applications motivate researchers to explore effective and efficient segmentation networks, particularly for mobile field.

\begin{figure}[t]
    \centering
    \includegraphics[width=.9\columnwidth]{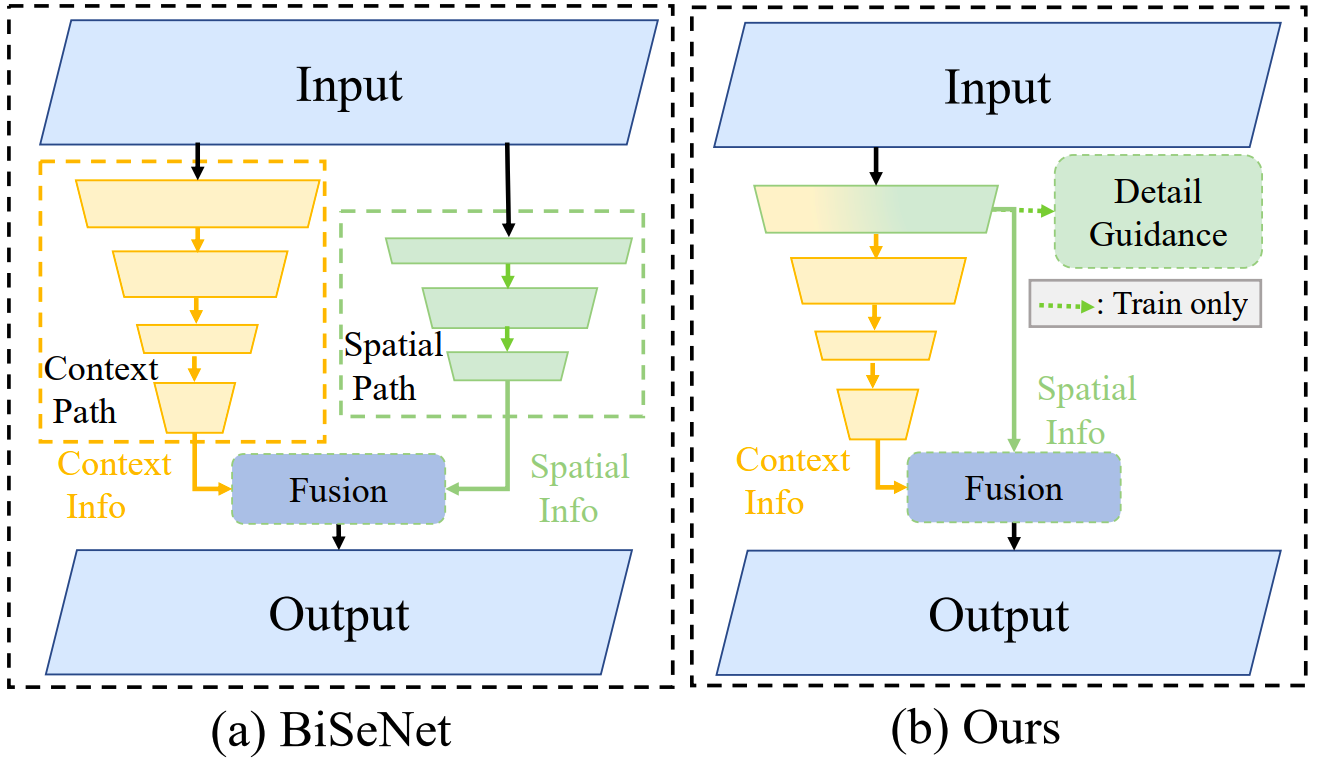} % Reduce the figure size so that it is slightly narrower than the column. Don't use precise values for figure width.This setup will avoid overfull boxes.
    \caption{Illustration of architectures of BiSeNet~\cite{Yu2018BiSenet} and our proposed approach. 
    (a) presents Bilateral Segmentation Network (BiSeNet~\cite{Yu2018BiSenet}), which use an extra \textit{Spatial Path} to encode spatial information. 
    (b) demonstrates our proposed method, which use a \textit{Detail Guidance} module to encode spatial information in the low-level features without an extra time-cosuming path.}
    \label{fig:architecture-comparison}
    \vspace{-0.5cm}
 \end{figure}

To fulfill those demands, many researchers propose to design low-latency, high-efficiency CNN models with satisfactory segmentation accuracy.
These real-time semantic segmentation methods have achieved promising performance on various benchmarks.
For real-time inference, some works, \textit{e.g.}, DFANet~\cite{li2019DFANet} and BiSeNetV1~\cite{Yu2018BiSenet} choose the lightweight backbones 
and investigate ways of feature fusion or aggregation modules to compensate for the drop of accuracy.
However, these lightweight backbones borrowed from image classification task may not be perfect for image segmentation problem due to the deficiency of task-specific design.
Besides the choice of lightweight backbones, restricting the input image size is another commonly used method to promote the inference speed.
Smaller input resolution seems to be effective, but it can easily neglect the detailed appearance around boundaries and small objects.
To tackle this problem, as shown in Figure~\ref{fig:architecture-comparison}(a), BiSeNet~\cite{Yu2018BiSenet, Yu2020BiSeNetV2} adopt multi-path framework to combine the low-level details and high-level semantics.
However, adding an additional path to get low-level features is time-consuming, and the auxiliary path is always lack of low-level information guidance.

To this end, we propose a novel hand-craft network for the purpose of faster inference speed, explainable structure, and competitive performance to that of existing methods. 
First, we design a novel structure, called Short-Term Dense Concatenate module (STDC module), to get variant scalable receptive fields with a few parameters.
Then, the STDC modules are seamlessly integrated into U-net architecture to form the STDC network, which greatly promote network performance in semantic segmentation task.

In details, as shown in Figure~\ref{backbone-architecture}, we concatenate response maps from multiple continuous layers, each of which encodes input image/feature in different scales and respective fields, leading to multi-scale feature representation.
To speed up, the filter size of layers is gradually reduced with negligible loss in segmentation performance. 
% our Short-Term Dense Concatenate Module (STDCM) can yield features with multi-scale information and sizeable receptive field.
% Besides, some basic modules, such as \textit{ConvX}, \textit{FC} layers, are leveraged to compose our STDC networks.
The details structure of STDC networks can be found in Table~\ref{table:network-settings}.
% And we stack the bottlenecks and other operations, such as \textit{ConvX}, \textit{FC} layers to form the STDC networks, as shown in Table~\ref{table:network-settings}. 

In the phase of decoding, as shown in Figure~\ref{fig:architecture-comparison}(b), instead of utilizing an extra time-consuming path, Detail Guidance are adopted to guide the low-level layers for the learning of spatial details.
We first utilize Detail Aggregation module to generate detail ground-truth.
Then, the binary cross-entropy loss and dice loss are employed to optimize the learning task of detail information, which is considered as one type of side-information learning.
It should be noted that this side-information is not required in the inference time.
Finally, the spatial details from low-level layers and semantic information from deep layers are fused to predict the semantic segmentation results.
% For decoder desgin, to further compansate the loss of low-level details due to the input restricting, we present an edge guidance training strategy which can be discarded in the inference phase.
% We add an extra auxiliary prediction heads named \textit{Detail Head} yield edge map of the images. 
% Then we adopt Laplacian operator to get the edge groundtruth from the segmantation groudtruth.
The whole architecture of our method is shown in Figure~\ref{fig:overall-architecture}.

Our main contributions can be summarized as follows:
\begin{itemize}
\vspace{-0.2cm}
\item We design a Short-Term Dense Concatenate module (STDC module) to extract deep features with scalable receptive field and multi-scale information. This module promotes the performance of our STDC network with affordable computational cost.
\vspace{-0.2cm}
\item We propose the Detail Aggregation module to learn the decoder, leading to more precise preservation of spatial details in low-level layers without extra computation cost in the inference time.
\vspace{-0.2cm}
\item We conduct extensive experiments to present the effectiveness of our methods. 
The experiment results present that STDC networks achieve new state-of-the-art results on ImageNet, Cityscapes and CamVid.
Specifically, our STDC1-Seg50 achieves  71.9\% mIoU on the Cityscapes \textit{test} set at a speed of \textbf{250.4 FPS} on one NVIDIA GTX 1080Ti card.
Under the same experiment setting, our STDC2-Seg75 achieves \textbf{76.8\% mIoU} at a speed of 97.0 FPS.
\vspace{-0.2cm}
\end{itemize}

%------------------------------------------------------------------------
\section{Related Work}

\subsection{Efficient Network Designs}
\noindent Model design plays an important role in computer vision tasks. 
% Running deep CNN models on embedded devices demands efficient model structures.
SqueezeNet~\cite{2016SqueezeNet} used the fire module and certain strategies to reduce the model parameters.
MobileNet V1~\cite{howard2017mobilenets} utilized depth-wise separable convolution to reduce the FLOPs in inference phase.
ResNet~\cite{he2016deep}~\cite{he2016identity} adopted residual building layers to achieve outstanding performance.
MobileNet V2~\cite{sandler2018mobilenetv2} and ShuffleNet~\cite{zhang2018shufflenet} used group convolution to reduce computation cost while maintaining comparable accuracy.
These works are particularly designed for the image classification tasks, and their extensions to semantic segmentation application should be carefully tuned.
\subsection{Generic Semantic Segmentation}
Traditional segmentation algorithms, \textit{e.g.}, threshold selection, super-pixel, 
utilized the hand-crafted features to assign pixel-level labels in images.
With the development of convolution neural network, methods~\cite{chen2017rethinking, badrinarayanan2017segnet, Zhao2017PSPnet, hu2020temporally} based on FCN~\cite{long2015fully} achieved impressive performance on various benchmarks.
The Deeplabv3~\cite{chen2017rethinking} adopted an atrous spatial pyramid pooling module to capture multi-scale context.
The SegNet~\cite{badrinarayanan2017segnet} utilized the encoder-decoder structure to recover the high-resolution feature maps.
The PSPNet~\cite{Zhao2017PSPnet} devised a pyramid pooling to capture both local and global context information on the dilation backbone.
% Hu~\textit{et al.}~\cite{hu2020temporally} reuse the high-level features of key frames to reduce computation cost.
Both dilation backbone and encoder-decoder structure can simultaneously learn the low-level details and high-level semantics.
However, most approaches require large computation cost due to the high-resolution feature and the complicate network connections.
In this paper, we propose an efficient and effective architecture which achieves good trade-off between speed and accuracy.
\subsection{Real-time Semantic Segmentation}
Recently, there are fast-growing practical applications for real-time semantic segmentation.
In this circumstance, there are two mainstreams to devise efficient segmentation methods.
(i) \textit{lightweight backbone.} 
DFANet~\cite{li2019DFANet} adopted a lightweight backbone to reduce computation cost and devised a cross-level feature aggregation module to enhance performance. 
DFNet~\cite{Li2019DFNet} utilized ``Partial Order Pruning'' algorithm to obtain a lightweight backbone and efficient decoder.
(ii) \textit{multi-branch architecture.} 
ICNet~\cite{zhao2018icnet} devised the multi-scale image cascade to achieve good speed-accuracy trade-off.
BiSeNetV1~\cite{Yu2018BiSenet} and BiSeNetV2~\cite{Yu2020BiSeNetV2} proposed two-stream paths for low-level details and high-level context information, separately. 
In this paper, we propose an efficient lightweight backbone to provide scalable receptive field.
Furthermore, we set a single path decoder which uses detail information guidance to learn the low-level details. 

\begin{figure}[t]
    \centering
    \includegraphics[width=1.0\columnwidth]{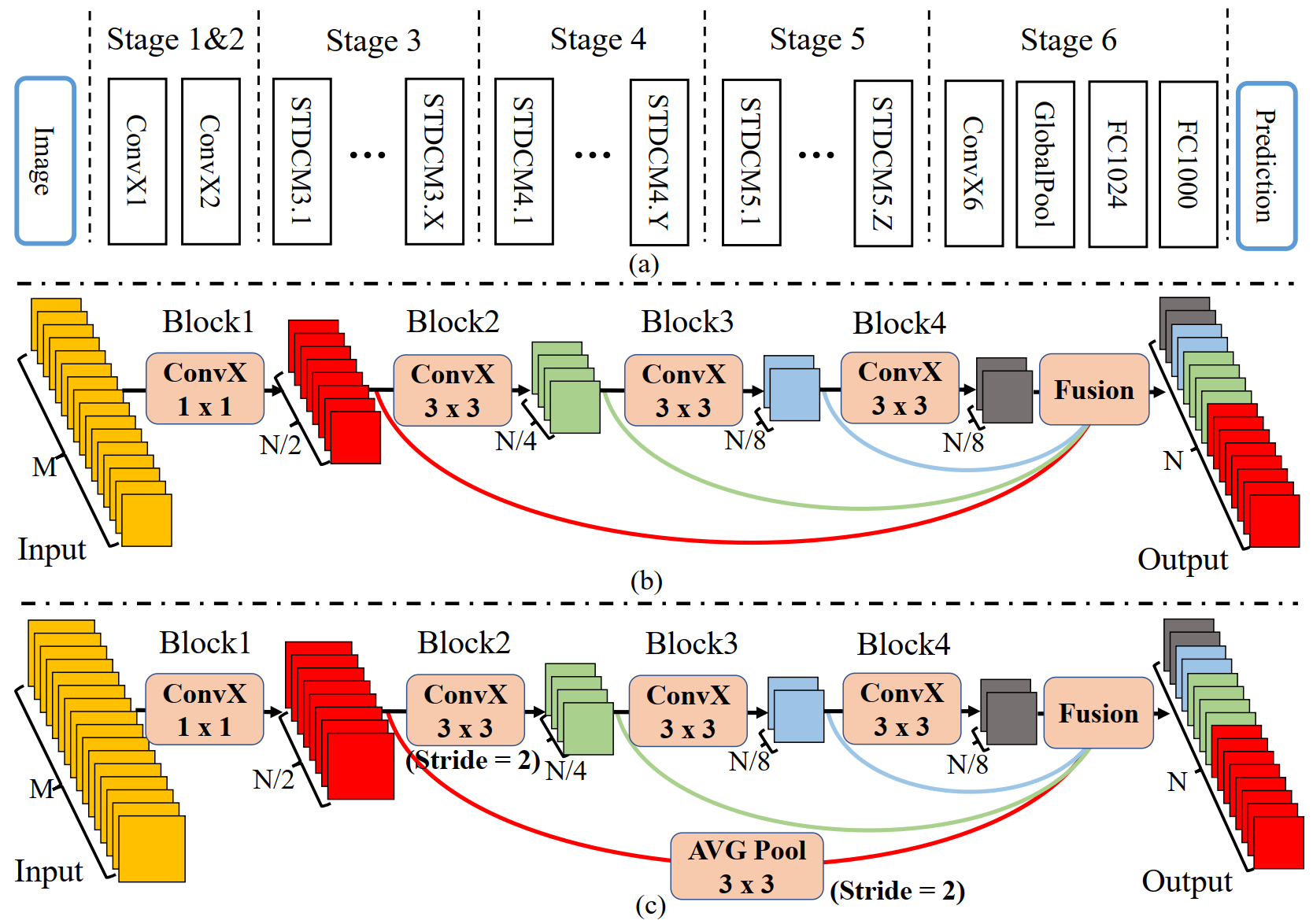} % Reduce the figure size so that it is slightly narrower than the column. Don't use precise values for figure width.This setup will avoid overfull boxes.
    \caption{
        (a) General STDC network architecture. \textit{ConvX} operation refers to the Conv-BN-ReLU.
        (b) Short-Term Dense Concatenate module (STDC module) used in our network. \textit{M} denotes the dimension of input channels, \textit{N} denotes the dimension of output channels. 
        Each block is a \textit{ConvX} operation with different kernel size.
        (c) STDC module with stride=2. 
    }
    \label{backbone-architecture}
    \vspace{-0.5cm}
\end{figure}
%-------------------------------------------------------------------------
\section{Proposed Method}
BiSeNetV1~\cite{Yu2018BiSenet} utilizes lightweight backbones, \textit{e.g.}, ResNet18 and spatial path as encoding networks to form two-steam segmentation architecture.
However, the classification backbones and two-stream architecture may be inefficient due to the structure redundancy.
In this section, we first introduce the details of our proposed STDC network. 
Then we present the whole arhitecture of our single-stream method with detail guidance.

\subsection{Design of Encoding Network}
% BiSeNetV1~\cite{Yu2018BiSenet} utilize lightweight backbones, \textit{e.g.}, ResNet18 and Xception39 as encoding network.
% These backbones may not be perfect for image segmentation due to the deficiency of task-specific information.
% In this section, we introduce the details of our proposed STDC network, which comprises of 6 stages, as shown in Figure~\ref{backbone-architecture}(a).

\subsubsection{Short-Term Dense Concatenate Module}

\begin{table}[t]
    \center
    \small 
    \setlength{\tabcolsep}{0.6mm}{
    \begin{tabular}{l|c|c|c|c|c}
    \Xhline{1.5pt}
    STDC module                  & Block1        & Block2       & Block3         & Block4           & Fusion  \\
    \hline RF(S = 1)       & $1 \times 1$  & $3 \times 3$ & $5 \times 5$   & $7 \times 7$     & \makecell[c]{$1 \times 1$, $3 \times 3$ \\ $5 \times 5$, $7 \times 7$} \\
    \hline RF(S = 2)       & $1 \times 1$  & $3 \times 3$ & $7 \times 7$   & $11 \times 11$   & \makecell[c]{$3 \times 3$ \\ $7 \times 7$, $11 \times 11$} \\

    \Xhline{1.5pt}
    \end{tabular}}
    \caption{Receptive Field of blocks in our STDC module. 
    \textit{RF} denotes Receptive Field, \textit{S} means stride,
    Note that if stride=2, the $1 \times 1$ \textit{RF} of Block1 is turned into $3 \times 3$ \textit{RF} by Average Pool operation.
    }
    \label{tb:blocks_receiptive_filed}
    \vspace{-.3cm}
\end{table}
% \textbf{Short-Term Dense Concatenate Module.} 
% \noindent We propose a efficient and effective module, named Short-Term Dense Concatenate Module (STDCM), to extract the response map with scalable receptive field in some layers of our network.
\noindent The key component of our proposed network is the Short-Term Dense Concatenate module (STDC module).
Figure~\ref{backbone-architecture}(b) and (c) illustrate the layout of STDC module.
Specifically, each module is separated into several blocks, and we use $ConvX_i$ to denote the operations of $i$-th block. Therefore, the output of $i$-th block is calculated as follows:
% In the first block, the input feature is processed by $1 \times 1$ kernel convolution as follows:
\begin{equation}
    x_{i} = ConvX_i(x_{i-1}, k_i) \label{eq:STDCM_conv}
\end{equation}
where $x_{i-1}$ and $x_i$ are the input and output of $i$-th block, separately.
ConvX includes one convolutional layer,  one batch normalization layer and ReLU activation layer, and $k_i$ is the kernel size of convolutional layer.

In STDC module, the kernel size of first block is 1, and the rest of them are simply set as 3.
Given the channel number of STDC module's output $N$, the filter number of convolutional layer in $i$-th block is $N/2^i$, except the filters of last convolutional layer, whose number is the same to that of previous convolutional layer. 
In image classification tasks, its a common practice to using more channels in higher layers.
But in semantic segmentation tasks, we focus on scalable receptive field and multi-scale informations.
Low-level layers need enough channels to encode more fine-grained informations with small receptive field, while high-level layers with large receptive field focus more on high-level information induction, setting the same channel with low-level layers may cause information redundancy.
Down-sample is only happened in Block2.
To enrich the feature information,
we concatenate $x_{1}$ to $x_{n}$ feature maps as the output of STDC module by skip-path.
Before concatenation, the response maps of different blocks in STDC module is down-sampled to the same spatial size by average pooling operation with $3 \times 3$ pooling size, as shown in Figure~\ref{backbone-architecture}(c).
In our setting, the final output of STDC module is:
% \vspace{-0.5cm}
\begin{equation}
    x_{output} = F(x_1,x_2, ..., x_{n}) \label{eq:STDCM}
% \vspace{-0.5cm}
\end{equation}
where $x_{output}$ denotes the STDC module output, $F$ is the fusion operation in our method, while $x_1,x_2, ..., x_{n}$ are feature maps from all $n$ blocks. 
In the consideration of efficiency, we adopt concatenation as our fusion operation.
In our method, we use the STDC module in $4$ blocks.
% It should be noted that we may down-sampled the spatial size of $x_1$ by Convolutional layer $AP(x, 3)$ in the skip-path, where $3$ is the kernel size.

Table~\ref{tb:blocks_receiptive_filed} presents the receptive field of blocks in STDC module, and $x_{output}$ thus gathers multi-scale information from all blocks,
% Furthermore, these information provide sizeable receptive field for the next block.
We claim that our STDC module has two advantages: 
(1) we elaborately tune the filter size of blocks by gradually decreasing in geometric progression manner, leading to significant reduction in computation complexity.
(2) the final output of STDC module is concatenated from all blocks, which preserves scalable respective fields and multi-scale information.

Given the input channel dimension $M$ and output channel dimension $N$, the parameter number of STDC module is:
% \begin{equation}
\begin{align}
    \nonumber S_{param} & = M \times 1 \times 1 \times \frac{N}{2^1} + \sum_{i=2}^{n-1} \frac{N}{2^{i-1}} \times 3 \times 3 \times \frac{N}{2^{i}} + \\
    \nonumber & \frac{N}{2^{n-1}} \times 3 \times 3 \times \frac{N}{2^{n-1}} \\
    \nonumber & = \frac{NM}{2} + \frac{9N^2}{2^3} \times \sum_{i=0}^{n-3}\frac{1}{2^{2i}} + \frac{9N^2}{2^{2n - 2}}  \\
    & = \frac{NM}{2} + \frac{3N^2}{2} \times (1 + \frac{1}{2^{2n - 3}}) 
    \label{eq:STDCM_param}
\end{align}
% \end{equation}
% where $M$ denote the dimension of input channels, $N$ denote the dimension of output channels.
As shown in Equation \ref{eq:STDCM_param}, the parameter number of STDC module is dominated by the predefined input and output channel dimension, while the number of blocks has slight impact on the parameter size.
Particularly, if $n$ reaches the maximum limit, the parameter number of STDC module almost keeps constant, which is only defined by $M$ and $N$.
% This means STDCM have limit paremeters regradless how many blocks it has.
% The dense concatenation schema of STDCM reduce the computation cost, also have sizeable receptive field, which is essential for real-time segmentation. 
\vspace{-0.5cm}
\begin{table}[t]
    \centering
    \normalsize
    \setlength{\tabcolsep}{1.1mm}{
    \begin{tabular}{c|c|c|c|c|c|c|c}
    \Xhline{1.5pt}
      \multirow{2}{*}{Stages}&
      \multirow{2}{*}{Output size}&
      \multirow{2}{*}{KSize}&
    %   \multirow{2}{*}{Stride}&
      \multirow{2}{*}{S}&
      \multicolumn{2}{c|}{STDC1} &
      \multicolumn{2}{c}{STDC2}\\
      \cline{5-6}
      \cline{7-8}
        % & & & & ~Repeat~ & Channels & ~Repeat~ & Channels \\
        & & & & ~R~ & C & ~R~ & C \\
      \hline
      Image& 224$\times$224& & & & 3& & 3\\
      \hline
      ConvX1& 112$\times$112& 3$\times$3& 2& 1& 32& 1& 32\\
      \hline
      ConvX2& 56$\times$56& 3$\times$3& 2& 1& 64& 1& 64\\
      \hline
      \makecell[c]{Stage3}&
      \makecell[c]{28$\times$28\\ 28$\times$28}&
      \makecell[c]{}&
      \makecell[c]{2\\ 1}&
      \makecell[c]{1\\ 1}&
      \makecell[c]{256}&
      \makecell[c]{1\\ 3}&
      \makecell[c]{256}\\
      \hline
      \makecell[c]{Stage4}&
      \makecell[c]{14$\times$14\\ 14$\times$14}&
      \makecell[c]{}&
      \makecell[c]{2\\ 1}&
      \makecell[c]{1\\ 1}&
      \makecell[c]{512}&
      \makecell[c]{1\\ 4}&
      \makecell[c]{512}\\
      \hline
      \makecell*[c]{Stage5}&
      \makecell[c]{7$\times$7\\ 7$\times$7}&
      \makecell[c]{}&
      \makecell[c]{2\\ 1}&
      \makecell[c]{1\\ 1}&
      \makecell[c]{1024}&
      \makecell[c]{1\\ 2}&
      \makecell[c]{1024}\\
      \hline
      ConvX6& 7$\times$7& 1$\times$1& 1& 1& 1024& 1& 1024\\
      \hline
      GlobalPool& 1$\times$1& 7$\times$7& & & & & \\
      \hline
      FC1& & & &  & 1024& & 1024\\
      \hline
      FC2& & & &  & 1000& & 1000\\
      \Xhline{1.0pt}
      FLOPs& & & & \multicolumn{2}{c|}{813M} & \multicolumn{2}{c}{1446M}\\
      \cline{5-6} 
      \cline{7-8}
      \hline
      Params& & & & \multicolumn{2}{c|}{8.44M} & \multicolumn{2}{c}{12.47M}\\
      \cline{5-6} 
      \cline{7-8}
      \Xhline{1.5pt}
    \end{tabular}}
    \caption{Detailed architecture of STDC networks. 
    Note that \textit{ConvX} shown in the table refers to the Conv-BN-ReLU.
    The basic module of Stage 3, 4 and 5 is STDC module. 
    KSize mean kernel size. S, R, C denote stride, repeat times and output channels respectively.}
    \label{table:network-settings}
    \vspace{-0.5cm}
\end{table}

\subsubsection{Network Architecture}
\noindent We demonstrate our network architecture in Figure~\ref{backbone-architecture}(a).
It consists of 6 stages except input layer and prediction layer.
Generally, Stage 1$\sim$5 down-sample the spatial resolution of the input with a stride of 2, respectively, 
and the Stage 6 outputs the prediction logits by one ConvX, one global average pooling layer and two fully connected layer.

The Stage 1\&2 are usually regarded as low-level layers for appearance feature extraction. 
In pursuit of efficiency, we only use one convolutional block in each of Stage 1\&2, which is proved to be sufficient
according to our experiences.
The number of STDC module in Stage 3, 4, 5 is carefully tuned in our network.
% consists of X, Y, Z Short-Term Dense Concatenate Modules, where X, Y, Z are intergers, \textit{e.g.} $X, Y, Z \in  \mathbb{N}^+$.
Within those stages, the first STDC module in  each stage down-samples the spatial resolution with a stride of 2.
The following STDC modules in each stage keep the spatial resolution unchanged.
% Stage 6 have a \textit{ConvX} named last conv, a global average pooling block and a fully connected block.
% Therefore, the whole architecture of our networks can be encoded as shown in Figure~\ref{backbone-architecture}(a).

% We utilize $S_l$ to denote output feature map of each stage, where $l$ is the index of stage $S_l$.
We denote the output channel number of stage as $N_l$, where $l$ is the index of stage.
In practice, we empirically set $N_6$ as 1024, and carefully tune the channel number of rest stages, until reaching a good trade-off between accuracy and efficiency.
% The width(number of channels) of $i_{th}$ stage is double ${i + 1}_{th}$ stage, \textit{e.g.} $C_{i+1} = 2 * C_{i}$.
Since our network mainly consists of Short-Term Dense Concatenate modules, we call our network STDC network.
Table~\ref{table:network-settings} shows the detailed structure of our STDC networks.

\begin{figure*}[t]
    \centering
    \includegraphics[width=.9\textwidth]{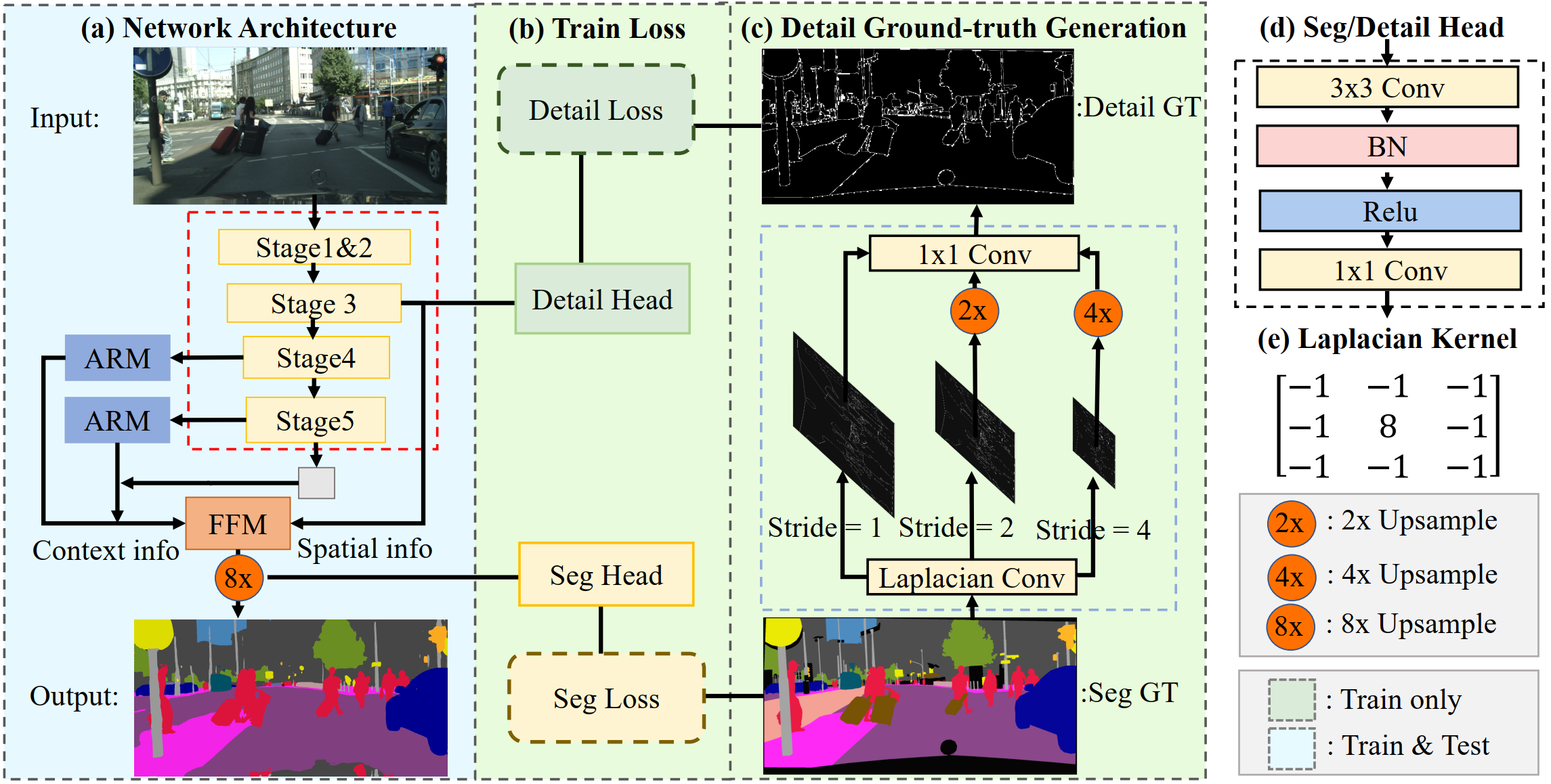} % Reduce the figure size so that it is slightly narrower than the column.
    \caption{Overview of the STDC Segmentation network. 
    \textit{ARM} denotes \textit{Attention Refine module}, 
    and \textit{FFM} denotes \textit{Feature Fusion Module} in ~\cite{Yu2018BiSenet}.
 %    \textit{Detail Head} and \textit{Seg Head} are Conv-BN-ReLU operators for the prediction of edge map and segmentation map.
    The operation in the dashed red box is our STDC network.
    The operation in the dashed blue box is \textit{Detail Aggregation Module}.
    % Note that the branches denoted by dotted line are only required in the training phase.
 %    For clarity, we draw the branch of \textit{Detail Head} in the Stage 3, which is actually parts of decoder as well.
    %is $3 \times 3$ Conv-BN-ReLU operator followed by one $1 \times 1$ convolution operation to get edge map and segmentation map, respectively. 
    }
    \label{fig:overall-architecture}
    \vspace{-0.2cm}
 \end{figure*}
%------------------------------------------------------------------------
\subsection{Design of Decoder}
% In this section, we introduce our segmentation architecture in details. 
% In general, our method abandon the time-consuming spatial path of BiSeNet~\cite{Yu2018BiSenet} and uses detail guidance to guide the architecture encode more spatial informations without extra computation cost.
% We first present the whole segmentation architecture of our method, then discuss the design of low-level features with detail guidance.
% \vspace{-0.5cm}
\subsubsection{Segmentation Architecture}
\noindent We use the pretrained STDC networks as the backbone of our encoder and adopt the context path of BiSeNet~\cite{Yu2018BiSenet} to encode the context information. 
As shown in Figure \ref{fig:overall-architecture}(a), we use the Stage 3, 4, 5 to produce the feature maps with down-sample ratio 1/8, 1/16, 1/32, respectively.
Then we use global average pooling to provide global context information with large receptive field.
The U-shape structure are deployed to up-sample the features stem from global feature, and combine each of them with the counterparts from last two stages~(Stage 4\&5) in our encoding phase.
Following BiSeNet~\cite{Yu2018BiSenet}, we use \textit{Attention Refine module} to refine the combination features of every two stages.
% For the final semantic segmentation prediction, we first concatenate the 1/8 down-sampled feature from the encoder and the counterpart from the decoder.
For the final semantic segmentation prediction, we adopt \textit{Feature Fusion module} in BiSeNet~\cite{Yu2018BiSenet} to fuse the 1/8 down-sampled feature from Stage 3 in the encoder and the counterpart from the decoder.
We claim that the features of these two stages are in different levels of feature representation. 
The feature from the encoding backbone preserves rich detail information, while the feature from the decoder contains context information due to the input from global pooling layer.
% Then, we use $1 \times 1$ Conv-BN-ReLU operator to reduce the channel dimension to 256,
% and use SE Attention Module in ~\cite{hu2018senet} to re-weight the feature.
% The re-weighted feature is then added to original feature to generate the final feature, which is used for the prediction of segmentation result by \textit{Seg Head}.
% We use seg head to generate the final results.
Specifically, the \textit{Seg Head} includes a $3 \times 3$ Conv-BN-ReLU operator followed with a $1 \times 1$ convolution to get the output dimension $N$, which is set as the number of classes.
% Moreover, two specific auxiliary \textit{Seg Heads} are leveraged to guide the output of intermediate layers in decoder.
We adopt cross-entry loss with Online Hard Example Mining to optimize the semantic segmentation learning task.

\begin{figure}[t]
    \centering
    \includegraphics[width=1.\columnwidth]{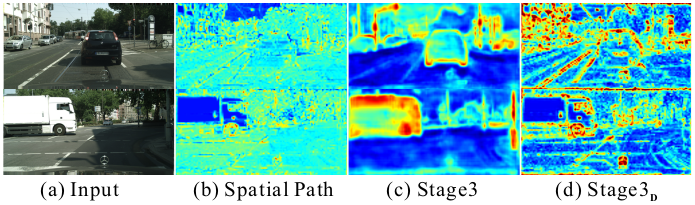} % Reduce the figure size so that it is slightly narrower than the column. Don't use precise values for figure width.This setup will avoid overfull boxes.
    \caption{Visual explanations for features in the spatial path and Stage 3 without and with \textit{Detail Guidance}.
    The column with subscript \textbf{D} denotes results with \textit{Detail Guidance}.
    The visualization shows that spatial path can encode more spatial detail,\textit{e.g.}, boundary, corners, than backbone's low-level layers, while our \textit{Detail Guidance} module can do the same thing without extra computation cost.
    }
    \label{fig:feature-visualization}
    \vspace{-0.6cm}
\end{figure}

\subsubsection{Detail Guidance of Low-level Features}

\noindent We visualize the features of BiSeNet's spatial path in Figure~\ref{fig:feature-visualization}(b).
Compared with the backbone's low-level layers(Stage 3) of same downsample ratio, spatial path can encode more spatial detail, \textit{e.g.}, boundary, corners.
Based on this observation, we propose a Detail Guidance module to guide the low-level layers to learn the spatial information in single-stream manner.
We model the detail prediction as a binary segmentation task.
We first generate the detail map ground-truth from the segmentation ground-truth by Laplacian operator as shown in Figure~\ref{fig:overall-architecture} (c).
% add an Detail Head to the low-level features from the backbone. 
As illustrated in Figure~\ref{fig:overall-architecture}(a), we insert the \textit{Detail Head} in Stage 3 to generate the detail feature map.
Then we use the detail ground-truth as the guidance of detail feature map to guide the low-level layers to learn the feature of spatial details.
As shown in Figure~\ref{fig:feature-visualization}(d), the feature map with detail guidance can encode more spatial details than aforementioned result presented in Figure~\ref{fig:feature-visualization}(c).
Finally, the learned detail features are fused with the context features from the deep block of the decoder for segmentation prediction.
% The ground-truth generation and loss function design are shown as following:
\\
\textbf{Detail Ground-truth Generation:} 
% \begin{figure}[t]
%     \centering
%     \includegraphics[width=1.\columnwidth]{edge-guidance} 
%     \caption{
%         Illustration of Detail GroundTruth Generation. \textit{2x, 4x} in the orange cycle denotes upsample operation. 
%         The operation in the dashed blue box is Detail Aggregation Module.
%     }
%     \label{fig:detail-guidance}
%    %  \vspace{-0.6cm}
% \end{figure}
We generate the binary detail ground-truth from the semantic segmentation ground-truth by our Detail Aggregation module, as shown in dashed blue box of Figure~\ref{fig:overall-architecture}(c).
This operation can be carried out by 2-D convolution kernel named Laplacian kernel and a trainable $1 \times 1$ convolution.
We use the Laplacian operator shown in Figure~\ref{fig:overall-architecture}(e) to produce soft thin detail feature maps with different strides to obtain mult-scale detail informations.
Then we upsample the detail feature maps to the original size and fuse it with a trainable $1 \times 1$ convolution for dynamic re-wegihting.
Finally, we adopt a threshold $0.1$ to convert the predicted details to the final binary detail ground-truth with boundary and corner informations.
% The Laplacian kernel is shown in Figure~ref{fig:} following:
% \begin{equation}
%     Laplacian~kernal =  \left[
%         \begin{array}{rcl}
%             -1&-1&-1 \\
%             -1&8&-1 \\
%             -1&-1&-1 \\
%         \end{array}
% \right]
% \label{eq:laplacian kernal}
% \end{equation}
% \begin{equation}
%     x_{output} =  [Conv() ,ConvX() ,ConvX()] \label{eq:STDCM}
% % \vspace{-0.5cm}
% \end{equation}
\\
\textbf{Detail Loss:} Since the number of detail pixels is much less than the non-detail pixels, detail prediction is a class-imbalance problem.
% Most related methods, \textit{edge detection}, often use weighted cross-entropy to solve this problem.
Because weighted cross-entropy always leads to coarse results,
following~\cite{deng2018crispBoundaries}, we adopt binary cross-entropy and dice loss to jointly optimize the detail learning.
Dice loss measures the overlap between predict maps and ground-truth.
Also, it is insensitive to the number of foreground/background pixels, which means it can alleviating the class-imbalance problem.
So for the predicted detail map with the height $H$ and the width $W$, the detail loss $L_{detail}$ is formulated as follows:
\begin{equation}
    L_{detail}(p_d, g_d) = L_{dice}(p_d, g_d) + L_{bce}(p_d, g_d)
\label{eq:edge_loss}
\end{equation}
where $p_d \in \mathbb{R}^{H \times W}$ denotes the predicted detail and $g_d \in \mathbb{R}^{H \times W}$ denotes the corresponding detail ground-truth.
$L_{bce}$ denotes the binary cross-entropy loss while $L_{dice}$ denotes the dice loss, which is given as follows:
\begin{equation}
    L_{dice}(p_d, g_d) = 1 - \frac{2\sum_{i}^{H \times W}p_d^ig_d^i + \epsilon}{\sum_{i}^{H \times W}(p_d^i)^2 + \sum_{i}^{H \times W}(g_d^i)^2 + \epsilon}
\label{eq:dice_loss}
\end{equation}
where $i$ denotes the $i$-th pixel and $\epsilon$ is a Laplace smoothing item to avoid zero division. In this paper we set $\epsilon = 1$.

As shown in Figure \ref{fig:overall-architecture}(b), we use a \textit{Detail Head} to produce the detail map, which guide the shallow layer to encode spatial information.
\textit{Detail Head} includes a $3 \times 3$ Conv-BN-ReLU operator followed with a $1 \times 1$ convolution to get the output detail map. %dimension N, here $N = 1$.
% Then we use a bilinear interpolation to up-sample the feature map to the size of ground-truth.
In the experiment, the \textit{Detail Head} is proved to be effective to enhance the feature representation.
Note that this branch is discarded in the inference phase.
Therefore, this side-information can easily boost the accuracy of segmentation task without any cost in inference.
% Therefore, it increases little computation cost in the inference phase and boost the accuracy in the task. 

\begin{figure*}[!t]
    \centering
    \includegraphics[width=0.98\textwidth]{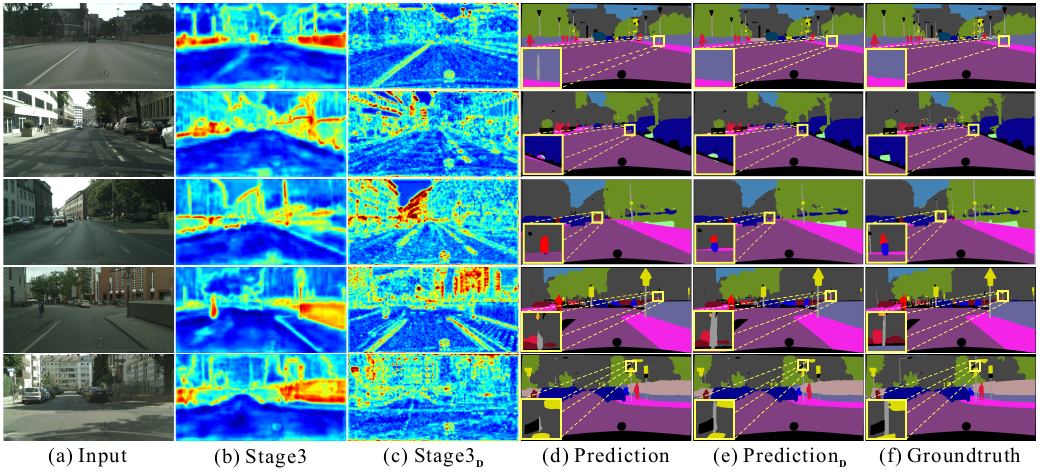} % Reduce the figure size so that it is slightly narrower than the column.
    \caption{Visual comparison of our \textit{Detail Guidance} on Cityscapes \textit{val} set. The column with subscript \textbf{D} denotes results with \textit{Detail Guidance}.
    The first row (a) shows the input images. (b) and (c) illustrate the heatmap of Stage 3 without and with \textit{Detail Guidance}.
    (d) and (e) demonstrate the predictions without and with \textit{Detail Guidance}. (f) is the ground-truth of input images. }
    \label{fig:edge-compare}
    \vspace{-0.2cm}
\end{figure*}

\begin{figure}[t]
    \centering
    \includegraphics[width=0.7\columnwidth]{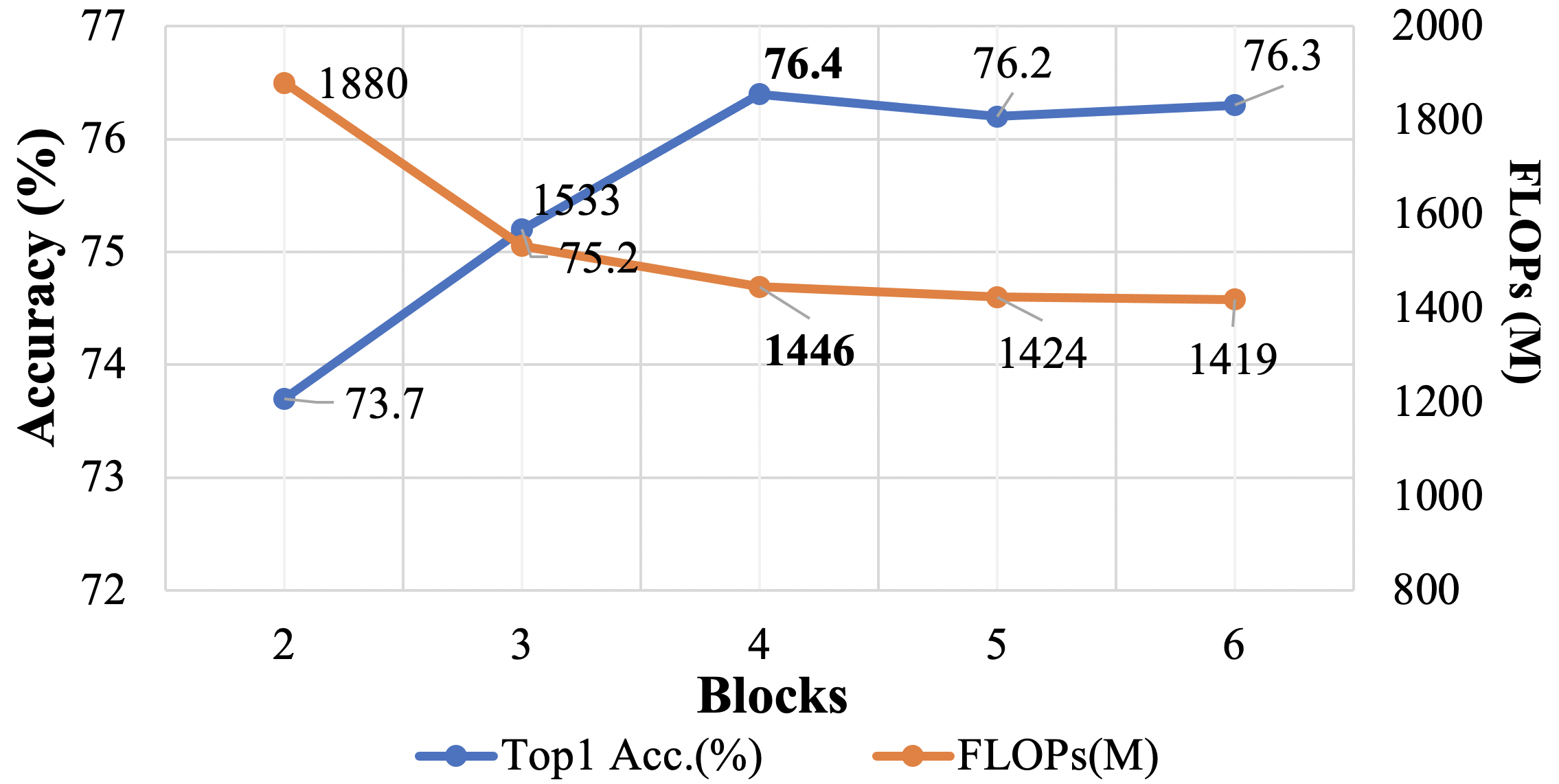} % Reduce the figure size so that it is slightly narrower than the column. Don't use precise values for figure width.This setup will avoid overfull boxes.
    \caption{Comparisons with different block number of STDC2 on ImageNet. }
    \label{fig:stdc_block_number}
    \vspace{-0.4cm}
\end{figure}

\vspace{-0.2cm}
\section{Experimental Results}
\noindent We implement our method on three datasets: ImageNet~\cite{deng2009imagenet}, Cityscapes~\cite{cordts2016cityscapes} and CamVid~\cite{BrostowSFC:ECCV08} to evaluate the effectiveness of our proposed backbone and segmentation network, respectively.
We first introduce the datasets and implementation details.
Then, we report our accuracy and speed results on different benchmarks compared with other algorithms. 
Finally, we discuss the impact of components in our proposed approach.
% \vspace{-0.4cm}
\subsection{Benchmarks and Evaluation Metrics}
% In order to verify the effectiveness and robustness of our method. \\
\noindent 
\textbf{ImageNet.} The ILSVRC~\cite{deng2009imagenet} 2012 is the most popular image classification dataset.
It contains 1.2 million images for training, and 50,000 for validation with 1,000 categories. It is also widely used for training a pretrained model for downstream tasks, like object detection or semantic segmentation.
\\
\textbf{Cityscapes.} Cityscapes~\cite{cordts2016cityscapes} is a semantic scene parsing dataset, which is taken from a car perspective.
It contains 5,000 fine annotated images and split into training, validation and test sets, with 2,975, 500 and 1,525 images respectively.
The annotation includes 30 classes, 19 of which are used for semantic segmentation task.
The images have a high resolution of $2,048 \times 1,024$, thus it is challenging for the real-time semantic segmentation.
For fair comparison, we only use the fine annotated images in our experiments.
\\
\textbf{CamVid.}
Cambridge-driving Labeled Video Database (Camvid)~\cite{BrostowSFC:ECCV08} is a road scene dataset, which is taken from a driving automobile perspective.
This dataset contains 701 annotated images extracted from the video sequence, in which 367 for training, 101 for validation and 233 for testing.
The images have a resolution of $960 \times 720$ and $32$ semantic categories, in which the subset of $11$ classes are used for segmentation experiments.
\\
\textbf{Evaluation Metrics.}
For classification evaluation, we use evaluate top-1 accuracy as the evaluation metrics following~\cite{he2016deep}. %and top-5 error
For segmentation evaluation, we adopt mean of class-wise intersection over union (mIoU) and Frames Per Second (FPS) as the evaluation metrics.
\vspace{-0.2cm}

\subsection{Implementation Details}
\noindent 
\textbf{Image Classification.}
We use mini-batch stochastic gradient descent (SGD) with batch size 64, momentum $0.9$ and weight decay $1e^{-4}$ to train the model.
Three training methods from \cite{zhang2019tricks} are adopted, including learning rate warmup, cosine learning rate policy and label smoothing. The total epochs is 300 with warmup strategy at the first 5 epochs, within which the learning rate starts from 0.001 to 0.1. The dropout before classification block is set to 0.2.
We do not use other special data augmentations, and all of them are the same as \cite{he2016deep}.
\\
\textbf{Semantic Segmentation.}
We use mini-batch stochastic gradient descent (SGD) with momentum $0.9$, weight decay $5e^{-4}$.
The batch size is set as $48$, $24$ for the Cityscapes, CamVid dataset respectively.
As common configuration, we utilize "poly" learning rate policy in which the initial rate is multiplied by $(1 - \frac{iter}{max\_iter})^{power}$. 
The power is set to $0.9$ and the initial learning rate is set as $0.01$.
Besides, we train the model for $60,000$, $10,000$ iterations for the Cityscapes, CamVid dataset respectively, in which we adopt warmup strategy at the first $1000$, $200$ iterations.

Data augmentation contains color jittering, random horizontal flip, random crop and random resize.
The scale ranges in [0.125, 1.5] and cropped resolution is $1024 \times 512$ for training Cityscapes. 
For training CamVid, the scale ranges in [0.5, 2.5] and cropped resolution is $960 \times 720$. 

In all experiments, we conduct our experiments base on pytorch-1.1 on a docker. 
We perform all experiments under CUDA 10.0, CUDNN 7.6.4 and TensorRT 5.0.1.5 on NVIDIA GTX 1080Ti GPU with batch size $1$ for benchmarking the computing power of our method. 
% \vspace{-0.2cm}
\subsection{Ablation Study}
\begin{table}[t]
    \center
    % \small 
    \setlength{\tabcolsep}{1.8mm}{
    \begin{tabular}{l|c|c|c}
    \Xhline{1.5pt}
    Backbone               & Resolution                                  &mIoU(\%)      & FPS \\
    \hline  GhostNet \cite{han2020ghostnet}       & $512 \times 1024$      & 67.8       & 135.0 \\
            MobileNetV3 \cite{howard2019searching}    & $512 \times 1024$      & 70.1   & 148.3  \\
            EfficientNet-B0 \cite{tan2019efficientnet} & $512 \times 1024$      & 72.2  & 99.9 \\
    \hline  STDC2           & $512 \times 1024$      & \textbf{74.2}                    & \textbf{188.6} \\
    \Xhline{1.0pt}
            GhostNet \cite{han2020ghostnet}       & $768 \times 1536$      & 71.3       & 60.9 \\
            MobileNetV3 \cite{howard2019searching}    & $768 \times 1536$      & 73.0   & 70.4 \\
            EfficientNet-B0 \cite{tan2019efficientnet} & $768 \times 1536$      & 73.9  & 45.9 \\
    \hline  STDC2           & $768 \times 1536$      & \textbf{77.0}                    & \textbf{97.0} \\

    \Xhline{1.5pt}
    \end{tabular}}
    \caption{Lightweight backbone comparison on Cityscapes \textit{val} set. 
    All experients utilize the same decoder and same experiment settings.}
    \label{tb:backbone_compare_results}
    \vspace{-0.5cm}
\end{table}

\begin{table}[t]
    \center
    % \small 
    \setlength{\tabcolsep}{1.6mm}{
    \begin{tabular}{l|c|ccc|c|c}
    \Xhline{1.5pt}
    \multirow{2}{*}{Method}&
    \multirow{2}{*}{SP}&
    % \multirow{2}{*}{Edge Guidance}&
    \multicolumn{3}{c|}{DG}&
    \multirow{2}{*}{mIoU(\%)}&
    \multirow{2}{*}{FPS}\\
%   \multirow{2}{*}{Stride}&
    % \multirow{2}{*}{S}&
    % \multicolumn{2}{c|}{STDC1} &
    % \multicolumn{2}{c}{STDC2}\\
    \cline{3-5}
    % \cline{7-8}
        % & & & & ~Repeat~ & Channels & ~Repeat~ & Channels \\
    & &4x &2x &1x & &  \\
    \hline
    BiSeNetV1~\cite{Yu2018BiSenet}&\checkmark  & & &  & 69.0 & 105.8  \\
    STDC2-50 &\checkmark & & &  & 73.7  & 171.6 \\
    \hline  STDC2-50 & & & &  & 73.0  & 188.6 \\
      STDC2-50 &  &\checkmark& &  & 73.4  & 188.6 \\
      STDC2-50 &  &&\checkmark&  & 73.6  & 188.6 \\
      STDC2-50 &  &&&\checkmark  & 73.8 & 188.6 \\
    \hline  STDC2-50 &  &&\checkmark&\checkmark  & 73.9  & 188.6 \\
      STDC2-50 &  &\checkmark&\checkmark&\checkmark  & \textbf{74.2}  & 188.6 \\

    \Xhline{1.5pt}
    \end{tabular}}
    \caption{Detail information comparison on Cityscapes \textit{val} set. 
    \textit{SP} means method with \textit{Spatial Path} 
    and \textit{DG} indicates \textit{Detail Guidance}, inwhich \textit{1x, 2x, 4x} denotes detail features with different down-sample strides in \textit{Detail Aggregation} module.}
    \label{tb:edge_compare_results}
    \vspace{-0.6cm}
\end{table}

\noindent This section introduces the ablation experiments to validate the effectiveness of each component in our method.
\\
\textbf{Effectiveness of STDC Module.}
% \begin{table}[h]
%     \center
%     \begin{tabular}{c|c|c|c|c}
%     \Xhline{1.5pt}
%     Block & Top1 Acc. & Params& FLOPs & depth \\
%     \hline 
%     2 & 72.7 & 12.57M & 1877M & 27 \\
%     3 & 74.2 & 10.36M & 1530M & 39 \\
%     4 & 75.4 & 9.80M & 1443M & 51 \\
%     5 & 75.2 & 9.66M & 1422M & 63 \\
%     6 & 75.3 & 9.63M & 1416M & 75 \\
%     \Xhline{1.5pt}
%     \end{tabular}
%     \caption{Comparisons with different block number of STDC2 on ImageNet.}
%     \label{tb:stdc_block_number}
% \end{table}
We adjust the block number of STDC module in STDC2 and present the result in Figure~\ref{fig:stdc_block_number}.
According to our Equation~\ref{eq:STDCM_param}, as the group number increases, the FLOPs decrease obviously.
And the best performance is in 4 blocks. 
The benefits of more blocks become very small and a deeper network is bad for the parallel calculation and FPS.
Hence, in this paper, we set the block number in STDC1 and STDC2 to 4.
\\
\textbf{Effectiveness of Our backbone.} To verify the effectiveness of our backbone designed for real-time segmentation, we adopt the latest lightweight backbones which has comparable classification performance compared with STDC2 to formulate a semantic segmentation network with our decoder.
As show in Table.~\ref{tb:backbone_compare_results}, our STDC2 yield the best speed-accuracy trade-off comapred with other lightweight backbones.
\\
\textbf{Effectiveness of Detail Guidance.} We first visualize the heatmap of the feature map of Stage $3$ as shown in Figure~\ref{fig:edge-compare}.
The features of Stage 3 with detail guidance encode more spatial information comparing to that of Stage $3$ without detail guidance.
Hence the final prediction of small objects and boundaries are more precise.
We show some quantitative results in Table~\ref{tb:edge_compare_results}.
To verify the effectiveness of our Detail Guidance, we show the comparison of different detail guidance strategies of STDC2-Seg on Cityscapes \textit{val} dataset.
To further demonstrate the capability of Detail Guidance, we first use the Spatial Path in BiSeNetV1~\cite{Yu2018BiSenet} to encode the spatial information,
then use the features generated from the Spatial Path to replace the features from Stage $3_D$.
The setting of experiment with Spatial Path are exactly the same with other experiments.
As shown in Table~\ref{tb:edge_compare_results}, Detail Guidance in STDC2-Seg can improve the mIoU without harming the inference speed.
Adding Spatial Path to encode spatial information can also improve the performance on accuracy, but it increases the computation cost at the same time.
Also we find our Detail Aggregation module encode the abundant detail information and yield the highest mIoU with aggregation of \textit{1x, 2x, 4x} detail features.
% Also we find that the higher image resolution is, the effectiveness of Edge Guidance is less.
% This may due to the reason that high resolution image contains enough edge information already.
% \vspace{-0.4cm}

\begin{table}[t]
    \small
    \setlength{\tabcolsep}{1.5mm}{
    \begin{tabular}{l|c|c|c|c}
    \Xhline{1.5pt}
     Model &  Top1 Acc.&  Params&  FLOPs & FPS  \\ %& RF \\
    \hline 
    ResNet-18 \cite{he2016deep} & 69.0\% & 11.2M & 1800M & 1058.7 \\ % & 435 \\
    ResNet-50 \cite{he2016deep} & 75.3\% & 23.5M & 3800M & 378.7 \\ % & 483 \\
    DF1 \cite{Li2019DFNet} & 69.8\% & 8.0M & 746M & 1281.3\\% &  \\
    DF2 \cite{Li2019DFNet} & 73.9\% & 17.5M & 1770M & 713.2\\% &  \\
    DenseNet121 \cite{huang2017densely} & 75.0\% & 9.9M & 2882M & 363.6\\% &  \\
    DenseNet161 \cite{huang2017densely} & 76.2\% & 28.6M & 7818M & 255.0\\% &  \\
    GhostNet(x1.0) \cite{han2020ghostnet} & 73.9\% & 5.2M & 141M & 699.1\\% &  \\
    GhostNet(x1.3) \cite{han2020ghostnet} & 75.7\% & 7.3M & 226M & 566.2\\% &  \\
    MobileNetV2 \cite{sandler2018mobilenetv2} & 72.0\% & 3.4M & 300M & 998.8\\% &  \\
    MobileNetV3 \cite{howard2019searching} & 75.2\% & 5.4M & 219M & 661.2\\% &  \\
    EfficientNet-B0 \cite{tan2019efficientnet} & 76.3\% & 5.3M & 390M & 443.0\\% &  \\
    \hline 
    STDC1 & 73.9\%       & 8.4M & 813M & \textbf{1289.0} \\% &  719\\
    STDC2 & \textbf{76.4}\% & 12.5M & 1446M & 813.6 \\% &  1002\\
    \Xhline{1.5pt}
    \end{tabular}}
    \caption{Comparisons with other popular networks on ImageNet Classification.}
    % \textit{RF} denotes receiptive field.
    \label{tb:classification_results}
    \vspace{-0.3cm}
\end{table}

% \vspace{-0.2cm}
\subsection{Compare with State-of-the-arts}
\noindent In this part, we compare our methods with other existing state-of-the-art methods on three benchmarks, ImageNet, Cityscapes and CamVid.
\\
\textbf{Results on ImageNet.} 
As shown in Table \ref{tb:classification_results}, our STDC networks achieves higher speed and accuracy compared with other lightweight backbones.
Compared with the lightweight backbones used in real-time segmentation, \textit{e.g.}, DF1Net, the top-1 classification accuracy of our STDC1 network is 4.1\% higher than baseline on the ImageNet \textit{validation} set.
Compared with populuar lightweight networks, \textit{e.g.} EfficientNet-B0, the FPS of STDC2 network is 83.7\% higher than that of baseline with competitive classification result.
\\
\begin{table}[t]
    \small 
    \setlength{\tabcolsep}{0.7mm}{
    \begin{tabular}{l|c|c|c|c|c}
    \Xhline{1.5pt}
    \multirow{2}{*}{Model} & \multirow{2}{*}{Resolution} & \multirow{2}{*}{Backbone} &  \multicolumn{2}{c|}{mIoU(\%)} & \multirow{2}{*}{FPS} \\
    \cline{4-5}
                                                  & & & val   & test &  \\
    \hline ENet~\cite{paszke2016enet} & $512 \times 1024$   &  no          & -     & 58.3 & 76.9 \\
        ICNet~\cite{zhao2018icnet}    & $1024 \times 2048$  &  PSPNet50    & -     & 69.5 & 30.3 \\
        DABNet~\cite{li2019dabnet}   & $1024 \times 2048$  &  no          & -     & 70.1 & 27.7 \\
        DFANet B~\cite{li2019DFANet} & $1024 \times 1024$  &  Xception B  & -     & 67.1 & 120  \\
        DFANet A'~\cite{li2019DFANet}& $512 \times 1024$   &  Xception A  & -     & 70.3 & 160  \\
        DFANet A~\cite{li2019DFANet} & $1024 \times 1024$  &  Xception A  & -     & 71.3 & 100  \\
        BiSeNetV1~\cite{Yu2018BiSenet}& $768 \times 1536$   &  Xception39  & 69.0  & 68.4 & 105.8  \\
        BiSeNetV1~\cite{Yu2018BiSenet}& $768 \times 1536$   &  ResNet18    & 74.8  & 74.7 & 65.5  \\
        CAS~\cite{zhang2019CAS}       & $768 \times 1536$  &  no          & -     & 70.5 & 108.0   \\
        GAS~\cite{Lin2020GAS}         & $769 \times 1537$  &  no          & -     & 71.8 & 108.4  \\
        DF1-Seg-d8~\cite{Li2019DFNet}& $1024 \times 2048$ &  DF1         & 72.4  & 71.4 & 136.9  \\
        DF1-Seg\cite{Li2019DFNet}  & $1024 \times 2048$  &  DF1         & 74.1  & 73.0 & 106.4  \\
        DF2-Seg1\cite{Li2019DFNet} & $1024 \times 2048$  &  DF2         & 75.9  & 74.8 & 67.2  \\
        DF2-Seg2\cite{Li2019DFNet} & $1024 \times 2048$  &  DF2         & 76.9  & 75.3 & 56.3  \\
        SFNet~\cite{li2020semantic}   & $1024 \times 2048$   &  DF1             & -  & 74.5 & 121  \\
        HMSeg~\cite{li2020humans}     & $768 \times 1536$    &  no             & -  & 74.3 & 83.2  \\
        TinyHMSeg~\cite{li2020humans} & $768 \times 1536$  &  no             & -  & 71.4 & 172.4  \\
        BiSeNetV2~\cite{Yu2020BiSeNetV2} & $512 \times 1024$  &  no          & 73.4  & 72.6 & 156  \\
        BiSeNetV2-L~\cite{Yu2020BiSeNetV2}& $512 \times 1024$  &  no         & 75.8  & 75.3 & 47.3  \\
        FasterSeg~\cite{chen2020fasterseg} & $1024 \times 2048$  &  no         & 73.1  & 71.5 & 163.9  \\

    \hline STDC1-Seg50 & $512 \times 1024$ &  STDC1     & 72.2  & 71.9 & \textbf{250.4}  \\
           STDC2-Seg50 & $512 \times 1024$ &  STDC2     & 74.2  & 73.4 & 188.6  \\
           STDC1-Seg75 & $768 \times 1536$ &  STDC1     & 74.5  & 75.3 & 126.7  \\
           STDC2-Seg75 & $768 \times 1536$ &  STDC2     & \textbf{77.0}  & \textbf{76.8} & 97.0  \\
    \Xhline{1.5pt}
    \end{tabular}}
    \caption{Comparisons with other state-of-the-art methods on Cityscapes. \textit{no} indicates the method do not have a backbone.}
    \label{tb:cityscapes_results}
    \vspace{-0.5cm}
\end{table}
\textbf{Results on Cityscapes.}
As shown in Table \ref{tb:cityscapes_results}, we present the segmentation accuracy and inference speed of our proposed method on Cityscapes \textit{validation} and \textit{test} set.
Following the previous methods~\cite{{Yu2020BiSeNetV2},{Lin2020GAS}}, we use the training set and validation set to train our models before submitting to Cityscapes online server.
At test phase, we first resize the image into the fixed size $512 \times 1024$ or $768 \times 1536$ to inference, then we up-sample the results to $1024 \times 2048$.
Overall, our methods get the best speed-accuracy trade-off among all methods.
We use $50$ and $75$ after the method name to represent the input size $512 \times 1024$ and $768 \times 1536$ respectively.
For example, with the STDC1 backbone and $512 \times 1024$ input size, we name the method STDC1-Seg50.
As shown in Table \ref{tb:cityscapes_results}, our STDC1-Seg50 achieves a significantly faster speed than baselines, \textit{i.e.}, \textbf{250.4 FPS}, and still has 71.9\% mIoU on \textit{test} set, 
which is over \textbf{45.2\%} faster than the runner-up.
Our STDC2-Seg50 using $512 \times 1024$ input size achieves 73.4\% mIOU with 188.6 FPS, which is the state-of-the-art trade-off between performance and speed.
For $768 \times 1536$ input size, our STDC2-Seg75 achieves the best mIOU 77.0\% in \textit{validation} set and \textbf{76.8\%} on \textit{test} set at 97.0 FPS.
\\
\textbf{Results on CamVid.}
We also evaluate our method on CamVid dataset. Table \ref{tb:comvid_results} shows the comparison results with other methods.
With the input size $720 \times 960$, STDC1-Seg achieves 73.0\% mIoU with 197.6 FPS which is the state-of-the-art trade-off between performance and speed.
This further demonstrates the superior capability of our method.
\begin{table}[t]
    \small 
    \setlength{\tabcolsep}{1.1mm}{
    \begin{tabular}{l|c|c|c|c}
    \Xhline{1.5pt}
    Model & Resolution & Backbone &  mIoU(\%) & FPS \\
    \hline ENet~\cite{paszke2016enet}     & $720 \times 960$    &  no          & 51.3  & 61.2 \\
           ICNet~\cite{zhao2018icnet}    & $720 \times 960$    &  PSPNet50    & 67.1  & 34.5 \\
           BiSeNetV1~\cite{Yu2018BiSenet}& $720 \times 960$    &  Xception39  & 65.6  & 175  \\
           BiSeNetV1~\cite{Yu2018BiSenet}& $720 \times 960$    &  ResNet18    & 68.7  & 116.3  \\
           CAS~\cite{zhang2019CAS}      & $720 \times 960$    &  no          & 71.2  & 169  \\
           GAS~\cite{Lin2020GAS}      & $720 \times 960$    &  no          & 72.8  & 153.1  \\
           BiSeNetV2~\cite{Yu2020BiSeNetV2} & $720 \times 960$   &  no          & 72.4  & 124.5  \\
           BiSeNetV2-L~\cite{Yu2020BiSeNetV2}& $720 \times 960$  &  no          & 73.2  & 32.7  \\
    \hline STDC1-Seg & $720 \times 960$ &  STDC1     & 73.0 & \textbf{197.6}  \\
           STDC2-Seg & $720 \times 960$ &  STDC2     & \textbf{73.9}  & 152.2  \\

    \Xhline{1.5pt}
    \end{tabular}}
    \caption{Comparisons with other state-of-the-art methods on CamVid. \textit{no} indicates the method do not have a backbone.}
    \label{tb:comvid_results}
    \vspace{-0.3cm}
\end{table}
\vspace{-0.6cm}
\section{Conclusions}
\noindent In this paper, we revisit the classical segmentation architecture BiSeNet~\cite{Yu2018BiSenet,Yu2020BiSeNetV2} for structure optimization. 
% and find out conducting information decomposition and induction effeciently and effeciently is very essiential for real-time segmentation.
Generally, the classification backbone and extra spatial path of BiSeNet greatly hinder the inference efficiency.
Therefore, we propose a novel Short-Term Dense Concatenate Module to extract deep features with scalable receptive field and multi-scale information.
Based on this module, STDC networks are designed and achieve competitive accuracy with high FPS in image classification.% task.
Using STDC networks as backbone, our detail-guided STDC-Seg achieves state-of-the-art speed-accuracy trade-off in real-time semantic segmentation.
Extensive experiments and visualization results indicates the effectiveness of our proposed STDC-Seg networks.
In future, we extend our method by following directions: 
(i) the backbone will be validated in more tasks, \textit{e.g.}, object detection. 
(ii) we will explore deeper on the utilization of spatial boundary in semantic segmentation tasks.

\noindent
\textbf{Acknowledgements}
This research is supported by Beijing Science and Technology Project. (No.Z181100008918018).

{\small
\bibliographystyle{ieee_fullname}
\bibliography{references}
}

\end{document}